\documentclass[runningheads]{llncs}
\usepackage{algorithm}
\usepackage{algpseudocode}
\usepackage[thinc]{esdiff}
\usepackage{graphicx}
\usepackage{caption}
\usepackage{subcaption}

\usepackage{tikz}
\usepackage{comment}
\usepackage{amsmath,amssymb} 
\usepackage{color}

\usepackage[accsupp]{axessibility}  

\usepackage[pagebackref=true,breaklinks=true,letterpaper=true,colorlinks,bookmarks=false]{hyperref}

\usepackage{booktabs}

\begin{document}
\pagestyle{headings}
\mainmatter
\def\ECCVSubNumber{3673}  

\title{Exploring Gradient-based Multi-directional Controls in GANs} 

\titlerunning{Exploring Gradient-based Multi-directional Controls in GANs}
%
\author{Zikun Chen\inst{1} \and
Ruowei Jiang \inst{1} \and
Brendan Duke\inst{1,2} \and
Han Zhao\inst{3} \and Parham Aarabi \inst{1,2}}
\authorrunning{Z. Chen et al.}
%
\institute{ModiFace Inc., \email{\{zikun,irene,brendan,parham\}@modiface.com}
 \and
University of Toronto
 \and
University of Illinois at Urbana-Champaign, \email{hanzhao@illinois.edu}
}
\maketitle

\begin{abstract}
Generative Adversarial Networks (GANs) have been widely applied in modeling diverse image distributions.
 However, despite its impressive applications, the structure of the latent space in GANs largely remains as a black-box, leaving its controllable generation an open problem, especially when spurious correlations between different semantic attributes exist in the image distributions.
 To address this problem, previous methods typically learn linear directions or individual channels that control semantic attributes in the image space. However, they often suffer from imperfect disentanglement, or are unable to obtain multi-directional controls.
  In this work, in light of the above challenges, we propose a novel approach that discovers nonlinear controls, which enables multi-directional manipulation as well as effective disentanglement, based on gradient information in the learned GAN latent space.
  More specifically, we first learn interpolation directions by following the gradients from classification networks trained separately on the attributes, and then navigate the latent space by exclusively controlling channels activated for the target attribute in the learned directions.
Empirically, with small training data, our approach is able to gain fine-grained controls over a diverse set of bi-directional and multi-directional attributes, and we showcase its ability to achieve disentanglement significantly better than state-of-the-art methods both qualitatively and quantitatively. 
The source code is available at \href{https://github.com/zikuncshelly/GradCtrl}{https://github.com/zikuncshelly/GradCtrl}.

\end{abstract}

\section{Introduction}

\begin{figure}[t]
\includegraphics[width=1.0\textwidth]{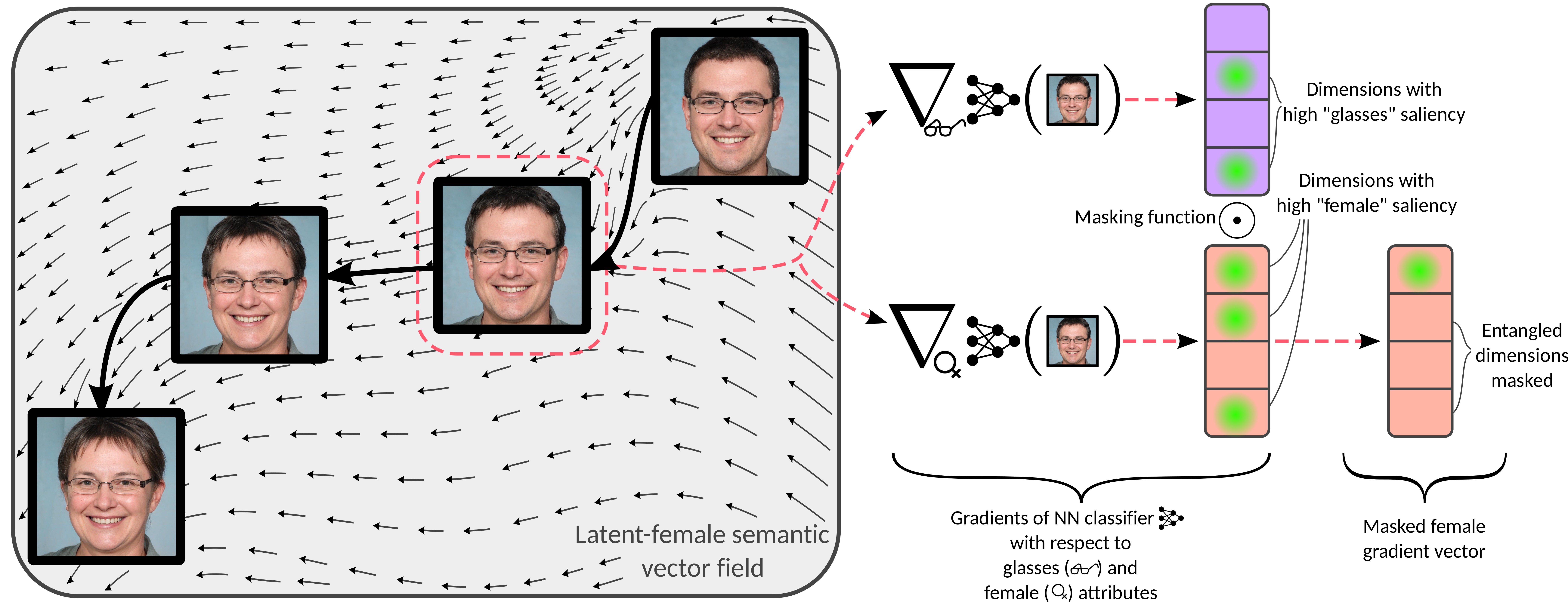}
\caption{Our method's semantic interpolation in a nonlinear vector field.
Directions corresponding to the ``female'' semantic in GAN latent space (left) are determined by differentiating a simple neural network classifier (middle).
Disentanglement is achieved by masking channels of the ``female'' semantic gradient vector that are highly salient for non-target attributes such as glasses (right).}
\label{fig:gradgan-overview}

\end{figure}
Generative adversarial networks (GANs)~\cite{goodfellow2014generative} are implicit generative models from which it is straightforward to sample data from the learned distributions. GANs work by learning to generate data from random noise sampled from natural distributions such as isotropic Gaussians. Due to the flexibility of neural networks in modeling complex transformations, GANs and the follow-up variants~\cite{brock2018large,karras2020analyzing,park2019semantic,wang2018high} have been widely applied in modeling image distributions.
Furthermore, the remarkable success of GANs' ability to generate realistically looking images has powered many real-world applications, including image inpainting~\cite{yeh2017semantic,yu2019free}, super-resolution~\cite{chen2021learning,ledig2017photo}, and image-to-image translations~\cite{Isola_2017_CVPR,zhu2017unpaired}.

However, despite the impressive existing applications, the structure of the latent space in GANs largely remains as a black box. For example, is there a correspondence between interpolation directions in the latent space and visual factors in the generated images? If yes, how can we generate images in a controllable way? Furthermore, are different interpolation directions independent of each other? If not, is it possible to disentangle them in an interpretable fashion?

To answer the above important but challenging questions, existing works have studied the learned latent space by identifying and linearly interpolating along directions that correspond to certain semantic factors in the generated images~\cite{harkonen2020ganspace,radford2015unsupervised,shen2020interfacegan,upchurch2017deep,wu2021stylespace}. For example, on facial image data, early works~\cite{radford2015unsupervised,upchurch2017deep} use simple linear arithmetic operations in the latent space to control the age of the generated persons. Recent works~\cite{shen2020interfacegan,yang2021semantic} further studied the interpolation directions by learning a linear binary classifier, and illustrated the effects of linearly interpolating different channels in a supervised or unsupervised manner~\cite{harkonen2020ganspace,wu2021stylespace}.

On the other hand, existing works aim to identify directions that correspond to unique semantic factors, and to control generation based on a single factor. However, semantic factors in images, such as age and eyeglasses, are often highly correlated with each other. As a result, artifacts often appear when applying controllable generation to one factor. Another issue arises due to spurious correlations in the training image data. These cause undesirable changes to the generated image to appear even if controllable generation is applied to an independent semantic factor. Furthermore, existing approaches still fail to discover multi-directional controls in the latent space and suffer from imperfect disentanglement between different interpolation directions. In fact, due to the nature of linear interpolation, existing works can only apply linear bi-directional controls over each semantic attribute. For example, to modify the color of a car, existing approaches need to learn all the pairwise connections among different colors, which is nontrivial~\cite{harkonen2020ganspace,wu2021stylespace}.


In light of the aforementioned challenges in controllable generations using GANs, in this work we propose a novel approach that first explores the multi-directional controls and addresses the disentanglement via gradient-based information in the learned GAN latent space (Fig.~\ref{fig:gradgan-overview}).
Inspired by existing visualization techniques on classification networks~\cite{chattopadhay2018grad,selvaraju2017grad}, we navigate the latent space by using gradient-based information learned from a classification network separately trained on the factors and attributes of interest.
Using the latent vectors from a pre-trained GAN network~\cite{karras2020analyzing} as inputs, we train classification networks over all attributes.
To disentangle the controllable generation of different attributes, we separate the manipulation into two steps.
First, we identify the direction using the gradient on the target attribute of interest.
Then, we exclusively control the channels that only affect the probability of the target attribute, based on gradient-weighted attribute activations.

To demonstrate the effectiveness of our approach, we present our method's ability to obtain nonlinear multi-directional controls as well as disentangled manipulations on multiple benchmark datasets.
Furthermore, we show that our classification network can gain proper controls by using a tiny dataset of 30 images per class.
We also compare our results to the state-of-the-art models both qualitatively and quantitatively.
Finally, we present real image manipulation results on several attributes by inverting the image back to a latent code.
To our knowledge, this work provides the first nonlinear multi-directional controls that achieve disentanglement successfully. Overall, our contributions are:
\begin{itemize}
	\item We propose a novel approach to navigate the GAN latent space by gradient-based information that allows nonlinear multi-directional controls.
	\item We disentangle the controls of different attributes by gradient-based channel selection to exclusively control a single attribute without affecting the others.
	\item Empirically, we show that our method discovers disentangled controls with small amounts of data, widening the range of applications of our method.
\end{itemize}

\section{Related Work}
\noindent
\textbf{Conditional Manipulations on GANs.}
GANs have been widely applied to a variety of tasks that allow conditional manipulations, including image editing~\cite{choi2018stargan,Isola_2017_CVPR,li2021continuous,liu2019stgan,park2019semantic,saha2021LOHO,shen2020interfacegan,wang2018high,zhu2017unpaired}, image inpainting~\cite{yeh2017semantic,yu2019free}, and super-resolution~\cite{chen2021learning,ledig2017photo}, just to name a few. Early works~\cite{Isola_2017_CVPR,zhu2017unpaired} explored image translations between two domains by using unpaired and paired training images.
Some more recent works focused on improving diversity by translating in multiple domains~\cite{choi2018stargan,choi2020stargan} while other works tried to improve the quality of the manipulation by effectively propagating the conditional information~\cite{park2019semantic}.
However, these methods often require a huge amount of paired data for training, which limits their applications in settings where it is expensive to manually label these pairs.

\noindent
\textbf{Understanding GAN Latent Space.} Many attempts have been made to understand and visualize the latent representations of GANs~\cite{bau2019gandissect,bau2019seeing,harkonen2020ganspace,radford2015unsupervised,shen2020interfacegan,upchurch2017deep,wu2021stylespace,yang2021semantic}.
Bau et al.~\cite{bau2019gandissect,bau2019seeing} used semantic segmentation models to detect and quantify the existence and absence of certain objects at both the individual and population levels.
Previous work~\cite{radford2015unsupervised,upchurch2017deep} empirically studied linear vector arithmetic in the latent representations.
To gain direct controls along attribute manipulations, later works~\cite{shen2020interfacegan,yang2021semantic} further explored linear interpolation in the latent space by finding semantically meaningful directions with explicit binary classifiers.
InterFaceGAN~\cite{shen2020interfacegan} hypothesized and showed the existence of a hyperplane serving as a linear separation boundary in the StyleGAN~\cite{karras2019style}  latent space for some binary semantic relationships.
In addition, they attempted to solve the entanglement issue on interpolated images via subspace projections between entangled attributes.
However, identity change is often observed in simple linear interpolations settings, and the disentangled direction sometimes fails to manipulate the target attribute.
A more recent work~\cite{wu2021stylespace} analyzed StyleGAN's style space and proposed a method to obtain valid semantic controls by identifying highly localized channels that only affect a specific region, which achieves well-disentangled changes on some attributes.
However, this approach lacks the ability to control globally aligned features such as age.

Another line of works aim to identify semantic controls in a self-supervised or unsupervised manner~\cite{cherepkov2021navigating,harkonen2020ganspace,voynov2020unsupervised}.
For example, GANSpace~\cite{harkonen2020ganspace} identifies important latent directions by applying principal component analysis (PCA) to vectors in GAN latent space or feature space. Voynov et al.~\cite{voynov2020unsupervised} discover interpretable directions in the GAN latent space by jointly optimizing the directions and a reconstructor that recovers these directions and manipulation strengths from images generated based on the manipulated GAN latent codes.
These types of methods usually involve extensive manual examinations of different manipulation directions as well as the identification of meaningful controls.
More importantly, in contrast to our approach, controlling over a target semantic is not always guaranteed and attributes are often observed as entangled due to the potential correlations in the training image distribution.

\noindent
\textbf{Gradient-based Network Understanding.} Gradient-based interpretation methods~\cite{chattopadhay2018grad,selvaraju2017grad,smilkov2017smoothgrad,sundararajan2017axiomatic} have been widely explored on discriminative models.
Grad-CAM~\cite{chattopadhay2018grad,selvaraju2017grad} used gradient information from the classification outputs to obtain localization maps that visualize evidence in images.
In contrast to local sensitivity, Integrated Gradients~\cite{sundararajan2017axiomatic} estimated the global importance of each pixel to address gradient saturation.
Integrated Gradients generated so-called gradient-based sensitivity maps, which were further sharpened by SmoothGrad~\cite{smilkov2017smoothgrad}.

The generative model LOGAN~\cite{wu2019logan} increased the stability of the training dynamics of GANs by adding an extra step before optimizing the generator and discriminator jointly.
In LOGAN, the latent code is first optimized towards regions regarded as more real by the discriminator, and the direction is obtained by calculating the gradient with respect to the latent codes.
Intuitively, the gradient points in a direction that obtains a better score from the discriminator.
We observed that when using LOGAN the discriminator gradient direction corresponds to meaningful transformations in GAN-synthesized images, for example, removal of artifacts, change in gender, and so on.
This observation inspired us to use auxiliary classifier gradients to obtain controls over GAN output semantics.
To the best of our knowledge, we are the first to take this approach.

\section{Methods}
In this section, we describe our proposed method to obtain disentangled controls in the GAN latent space.
Our method first finds semantically meaningful directions by calculating gradients of multi-class classifiers that score different semantics given latent code inputs.
Next, by selecting the channels in the latent code that correspond to these directions, our technique achieves attribute disentanglement during manipulation through masking.

\begin{figure}
\includegraphics[width=1\textwidth]{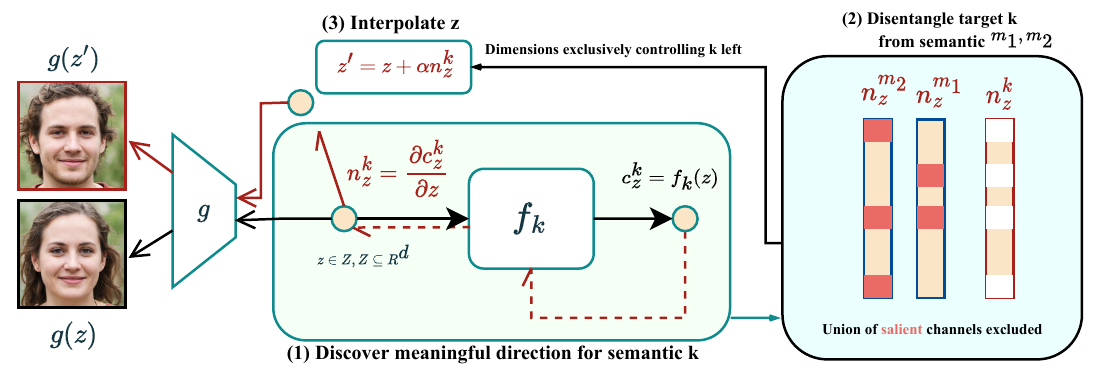}
\caption{Overview of our method. We first obtain the control for a semantic attribute $k$, e.g., gender, by calculating the gradient direction $n_{z}^{k}$ from the mapping function $f_k$ in (1). To disentangle direction $n_{z}^{k}$ from attributes $m_1, m_2$, e.g., smile and age, the union of salient dimensions in control directions $n_{z}^{m_{1}}, n_{z}^{m_{2}}$ for the corresponding semantics are masked to 0 in $n_{z}^{k}$, as shown in (2). We then interpolate the latent code following the direction computed at each step in (3).}

\end{figure}

\subsection{Discovery of Semantically Meaningful Directions}
\noindent
\textbf{Learning Semantics in GAN Latent Space.} GANs learn a mapping~$g : Z \rightarrow X$ from a latent code $z$ in its latent space $Z \subseteq \mathbb{R}^d$ to an image $x = g(z)$ in the image space $X\subseteq \mathbb{R}^{H\times W\times3}$, where~$H\times W$ is the spatial resolution.
Human-interpretable semantics such as age of a person exist in the image space.
Given a series of scoring functions $(s_1,\dots,s_K)$ for $K$ semantics, we can obtain $K$ mappings from the GAN latent space $Z$ to the semantic spaces $C_1,\dots,C_K$ respectively, where
\begin{equation}
c_{z}^{k} = s_k(g(z)) \in C_k.
\label{eq:latent-semantic-mapping}
\end{equation}
Here $c_{z}^{k}$ denotes the $k$th semantic score of the image generated based on the latent code $z$, and $C_k \in \mathbb{R}^{d_k}$, where $d_k$ denotes the number of classes the $k$th semantic has, e.g., for binary semantics, $d_k=1$. We parametrize each mapping function $s_k(g(z))$ with a neural network $f_k$ trained on paired samples of GAN latent codes and corresponding semantic labels.
The labels are generated by pre-trained image classifiers or human supervision.
With $(f_1,\dots,f_k)$, we are able to estimate human-interpretable semantics based on GAN latent codes, without inferring from the image space.

\noindent
\textbf{Manipulating Semantics in GAN Latent Space.}
Provided the accuracy of $(f_1,\dots,f_k)$, it is expected that when controlling the $k$th semantic by interpolating $z$ along meaningful directions, its score $c_{z}^{k}$ changes accordingly.
We hypothesize that such changes correspond to the changes in image space semantics. Inspired by LOGAN~\cite{wu2019logan}, which hypothesizes that the gradient from a GAN discriminator points in the direction of better samples, as the discriminator can be seen as a scoring function for ``realness'', we propose to control the $k$ semantics using the Jacobians of $(f_1,\dots,f_k)$. In particular, our method controls~$c_{z}^{k}$ by interpolating the latent code following gradient directions that are rows of the Jacobian of~$f_k$ with respect to $z$:
\begin{equation}
        n^{k}_{z} = \nabla f_k(z) = \diff{f_k(z)}{z} \in \mathbb{R}^{d_k \times d},
        \label{eq:latent-semantic-jacobian}
\end{equation}
where the matrix $n^{k}_{z}$ denotes the Jacobian of the~$k$th latent-semantic mapping~$c_{z}^{k}$ at $z$. Note that for binary attributes, $d_k=1$ and we use $n^{k}_{z}=n^{k}_{z}[0]$ to denote a gradient vector. For multi-class attributes, when optimizing towards the $j$th class for semantic  $k$, we take the corresponding row in the Jacobian and $n^{k}_{z}=n^{k}_{z}[j]$. For notational simplicity, $n^{k}_{z}[j]$ is abbreviated as $n^{k}_{z}$ in the following sections, so that~$n^k_z$ always refers to a gradient vector.

Intuitively, for each target semantic, interpolating the latent code following the gradient from its scoring function changes the semantic score, then the semantic is manipulated accordingly. More specifically, to manipulate the semantic $k$ with respect to its $i$th class once, we update $z$ with

\begin{equation}
z' = z + \alpha n^{k}_{z},
\label{eq:latent-interpolation}
\end{equation}
where $\alpha$ is a hyperparameter that controls the interpolation step size and sign of change in the target semantic value.
In contrast to works that discovered linear semantically meaningful directions applicable to all coordinates in GAN latent space~\cite{shen2020interfacegan}, our method finds unique local and data-dependent directions for each coordinate due to the nonlinear nature of the latent-semantic mapping (Eqn.~\ref{eq:latent-semantic-mapping}).




\subsection{Disentanglement of Attributes During Manipulation}
Entanglements such as changing the gender affecting one's age can emerge during latent code interpolation following the gradient of the latent-semantic mapping (Eqn.~\ref{eq:latent-interpolation}). In this section, we propose a technique to minimize this effect.

\noindent
\textbf{Target-specific Dimensions in Direction Vectors.} In early experiments, we observed that interpolation along the gradient direction alters non-target semantics, but disentanglement is occasionally achieved by randomly excluding dimensions in the direction vector used for interpolation.
Therefore, we hypothesize that within the $d$-dimensional direction vector discovered, only some dimensions are responsible for the change in the target semantic.
Conversely, the remaining channels denote bias learned from the training data.
For example, when interpolating in the direction that increases a person's age, eyeglasses appear since eyeglasses and age are correlated in the dataset.

\noindent
\textbf{Filtering Dimensions to Control Target Attributes Exclusively.} To construct disentangled controls, we propose to filter out dimensions based on gradient magnitudes from the semantic scoring function in a technique inspired by Grad-CAM~\cite{selvaraju2017grad}.
Grad-CAM measures the importance of neurons by
 \begin{equation}
 a^{c}_{k}= \frac{1}{Z}\sum_i \sum_j \frac{\partial y^c}{\partial A_{ij}^{k}},
 \label{eq:gradcam-importance}
\end{equation}
where the derivative of the score $y^c$ for class $c$ with respect to the activation map $A$ is average pooled over spatial dimensions.
The final heatmap is then
\begin{equation}
        L_{\mathrm{Grad-CAM}}^{c} = \mathrm{ReLU}\big(\sum_k a^{c}_{k} A^k \big).
\end{equation}
For our approach, gradient $n^{k}_{z}$ is regarded as the only activation map, and the saliency of the $i$th dimension is calculated by
\begin{equation}
L_{i}^{k} = |{(n^{k}_{z})}_i|,
\end{equation}
where $|(n^{k}_{z})_i|$ denotes the absolute value of the $i$th element of the gradient vector.
Contrary to the original classification setting for Grad-CAM where ReLU is chosen as the activation, here the absolute value is used because in our classification setting dimensions with negative influence on the semantic score also contain meaningful information.

\begin{algorithm}[tb]
\caption{Disentangle attributes}
\label{algo1}
\begin{algorithmic}[1]
    \State Target attribute $k$ is entangled with attributes $m_1,\dots,m_j$ in direction $n_{z}^{k}$
    \State Scoring function $f_{m_1},\dots,f_{m_j}$, numbers of top channels excluded $\{c_1,\dots,c_j\}, j\neq k$
    \State Excluded set of dimensions $E=\{\}$
    \State For each $(m,f_m,c) \in \mathcal (\{m_1,\dots,m_j\}$ ,$ \{f_{m_1},\dots,f_{m_j}\}$,$\{c_1,\dots,c_j\})$:
        \State \hskip2em $n^{m}_{z} = \nabla f_m(z) = \diff{f_m(z)}{z}$, $L_{i}^{m}=|{(n^{m}_{z})}_i|$
        \State \hskip2em $t = sorted(L^m)[-c]$, $e = \{i\}$ where $L_{i}^{m}\geq t$

        \State \hskip2em $E=E\cup e$
    \State ${{(n_{z}^{k})}}_i = 0$ for $i \in E$
    \State Return ${n_{z}^{k}}$
\end{algorithmic}
\end{algorithm}

By definition, the value of gradient $n^{k}_{z}$ at its $i$th dimension indicates the rate of change in $c_{z}^{k}$ induced by a small change in latent vector $z$ along dimension~$i$.
Intuitively, dimensions in $n^{k}_{z}$ with greater saliency $L_{i}^{k}$ are more relevant and have more impact on $c_{z}^{k}$, whereas less relevant dimensions for $c_{z}^{k}$ could have high saliency in another semantic $m$'s gradient $n^{m}_{z}, m \neq k$.
Hence even small changes in these dimensions while optimizing for the $k$th semantic could affect $c_{z}^{m}$ significantly, causing attribute entanglement.
Note that such effects could be either positive or negative, hence we use the absolute values as the saliency.
To alleviate this issue, we exclude the top-$k$ salient dimensions for predicting an entangled attribute from the target direction, with $k$ in \{50,100,150,200,250\}.
Algo.~\ref{algo1} describes the full process to calculate the new direction with disentangled attributes before interpolating the latent code (Eqn.~\ref{eq:latent-interpolation}).

\section{Experiments}
In this section, we apply our method to learned latent representations by state-of-the-art GAN models, and compare it with existing state-of-the-art latent code interpolation methods both qualitatively and quantitatively. In particular, we analyze the ability to achieve disentangled controls using each method.

\subsection{Qualitative Results}

\noindent
\textbf{Image Bank.} We perform our experiments on StyleGAN2~\cite{karras2020analyzing} pretrained on FFHQ~\cite{karras2019style}, LSUN Car~\cite{yu2015lsun}, LSUN Cat~\cite{yu2015lsun}, and StyleGAN~\cite{karras2019style} pretrained on LSUN Bedroom~\cite{yu2015lsun}.
To prepare an image bank, we synthesize images by randomly sampling the latent codes and obtain labelled pairs for each attribute value, via pre-trained classifiers or human supervision.

\noindent
\textbf{Training the Classifier.} We train a simple classifier with only 30 examples for each attribute value. The network consists of two fully connected layers, followed by the corresponding prediction head: sigmoid for binary classification and softmax for multi-class classification.
In the Appendix, we also demonstrate the effects of using different numbers of training samples.

\noindent
\textbf{Interpolating in the Latent Space.}
We interpolate the latent codes following the gradient directions of our trained classifiers by a fixed step size of 0.6.
We perform layer-wise edits by applying the learned directions to certain layers in \textbf{W+} space in StyleGAN/StyleGAN2.
More implementation details can be found in the Appendix.

\begin{figure}[tb]
\centering
\includegraphics[width=\textwidth]{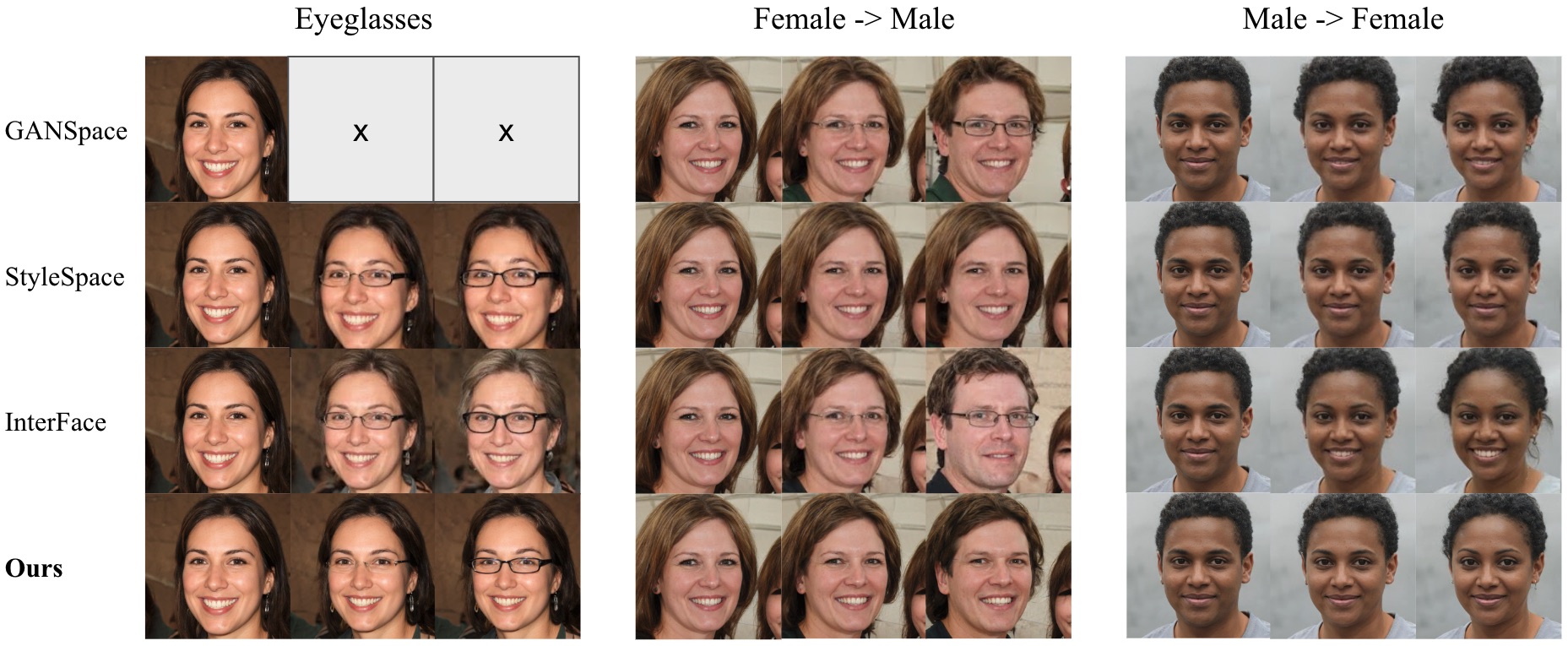}
\includegraphics[width=\textwidth]{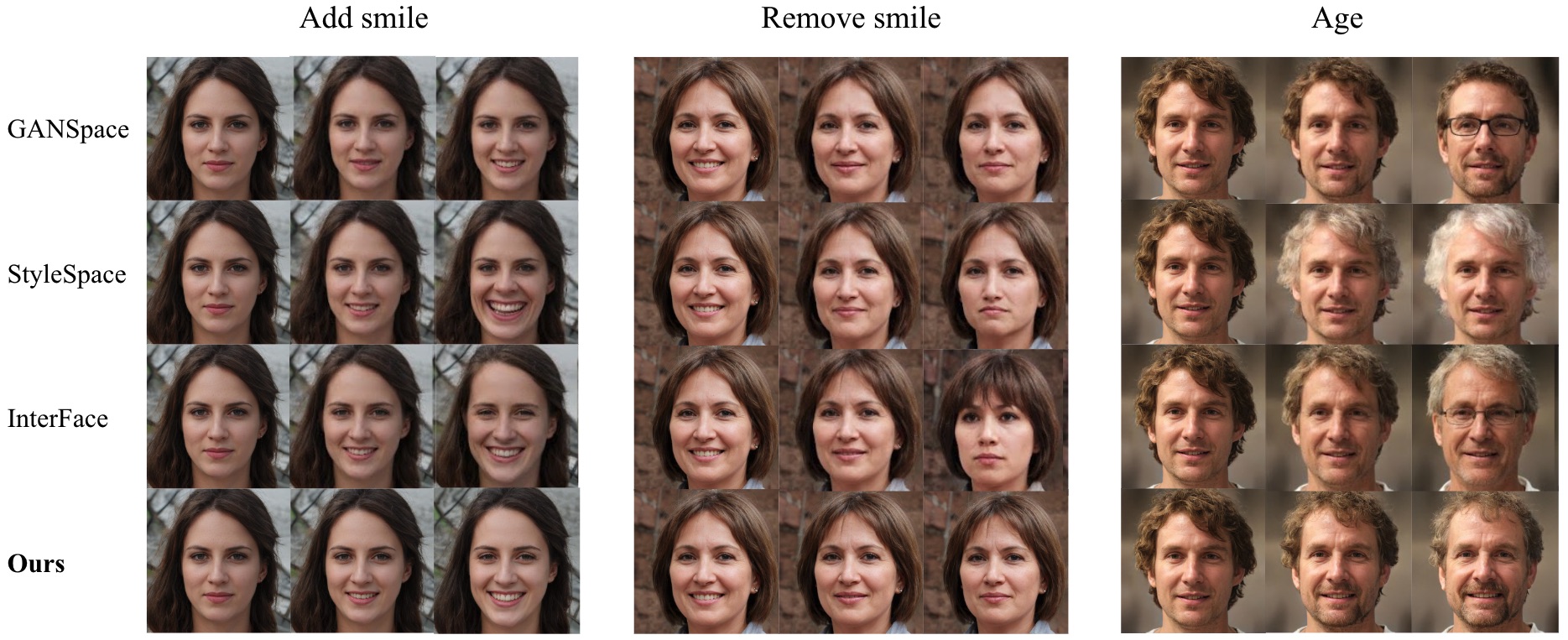}
\caption{Comparison with state-of-the-art methods on StyleGAN2~\cite{karras2020analyzing} pretrained on FFHQ~\cite{karras2019style}. For each group of images, left: source; middle, right: small and large step size. Our method edits eyeglasses and smiles naturally, and for gloabl-level attributes, ours is able to produce more visually distinguishable editing results compared to StyleSpace~\cite{wu2021stylespace} and GANSpace, and achieves more disentangled control than InterFaceGAN~\cite{shen2020interfacegan}.}
\label{ffhq_compare}
\end{figure}

\noindent
\textbf{Disentangled Attribute Manipulation.}
In this section, we evaluate our method on StyleGAN2 models pretrained on the FFHQ~\cite{karras2019style}, LSUN-Cars~\cite{yu2015lsun}, LSUN-Cats datasets~\cite{yu2015lsun} and the StyleGAN model~\cite{karras2019style} pretrained on the LSUN-Bedroom dataset~\cite{yu2015lsun}. We demonstrate that our approach obtains multi-directional controls over GAN-generated images in a disentangled manner.
We show a qualitative comparison among key attributes in the FFHQ dataset~\cite{karras2019style}: smile, gender, age, and eyeglasses (Fig.~\ref{ffhq_compare}).
For the age and gender attributes, not only is our approach able to generate realistic localized target effects such as wrinkles or acne for age, but it is also capable of editing the image on a global level while preserving the person’s identity and irrelevant attributes disentangled, such as keeping the person's smile when changing from female to male.

\begin{figure}[tb]
\centering
\includegraphics[width=\textwidth]{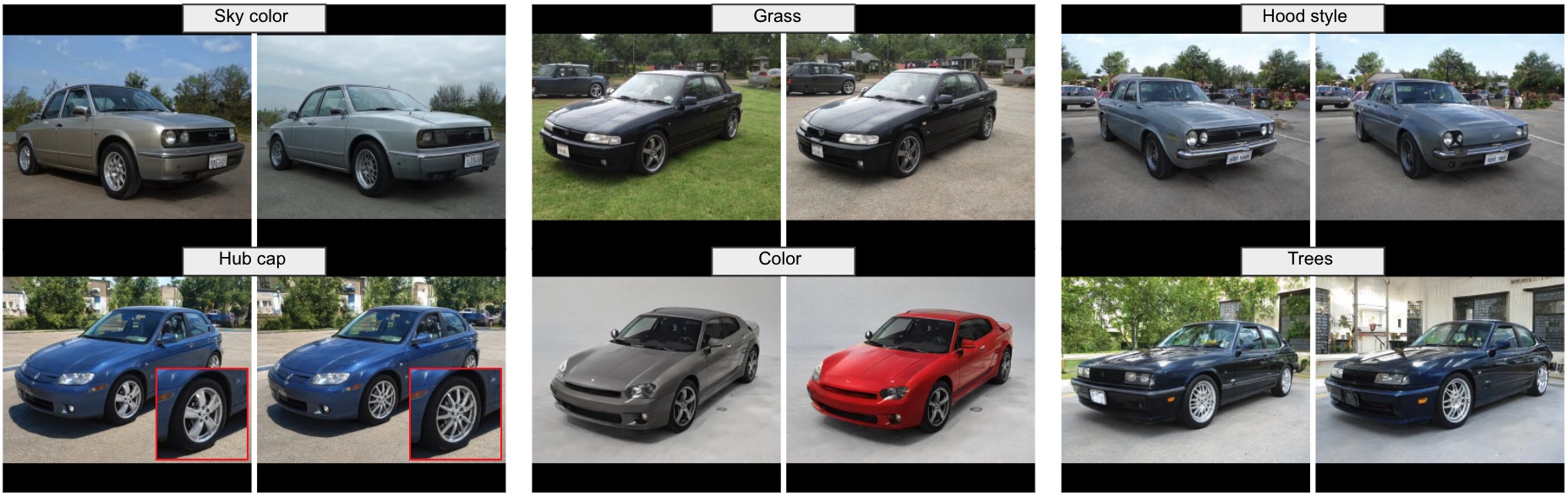}
\includegraphics[width=\textwidth]{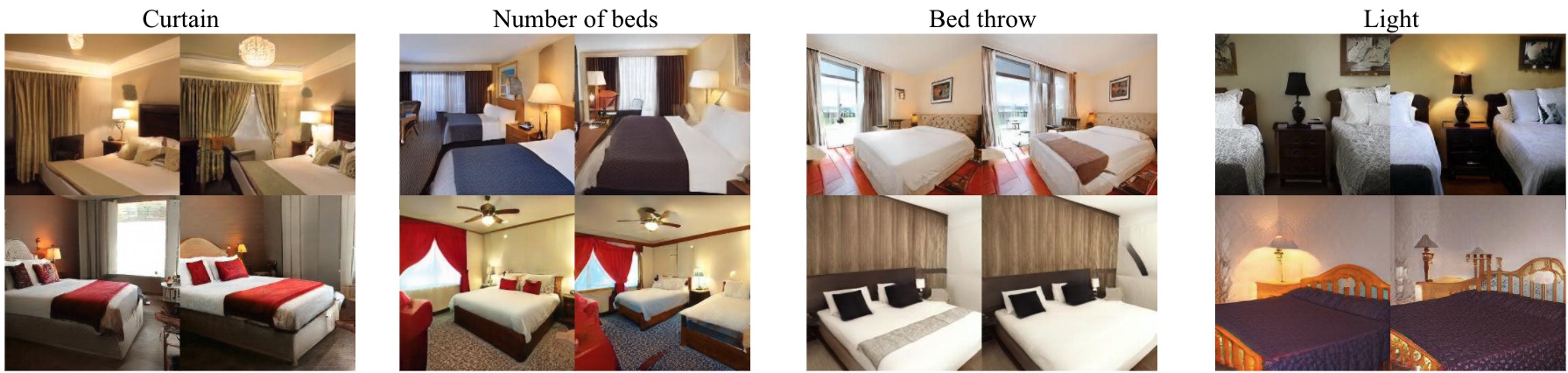}
\caption{LSUN-Cars and LSUN-Bedroom~\cite{yu2015lsun} editing results only trained on 30 pairs of examples. For each group of 4 images, left: source, right: modified.}
\label{lsun_examples}
\end{figure}

We further show that our method works on a variety of attributes on multiple datasets with qualitative results on LSUN car and LSUN bedroom~\cite{yu2015lsun} (Fig.~\ref{lsun_examples}).

\begin{figure}[tb]
\centering
\begin{subfigure}{1\textwidth}
\centering
\includegraphics[width=\textwidth]{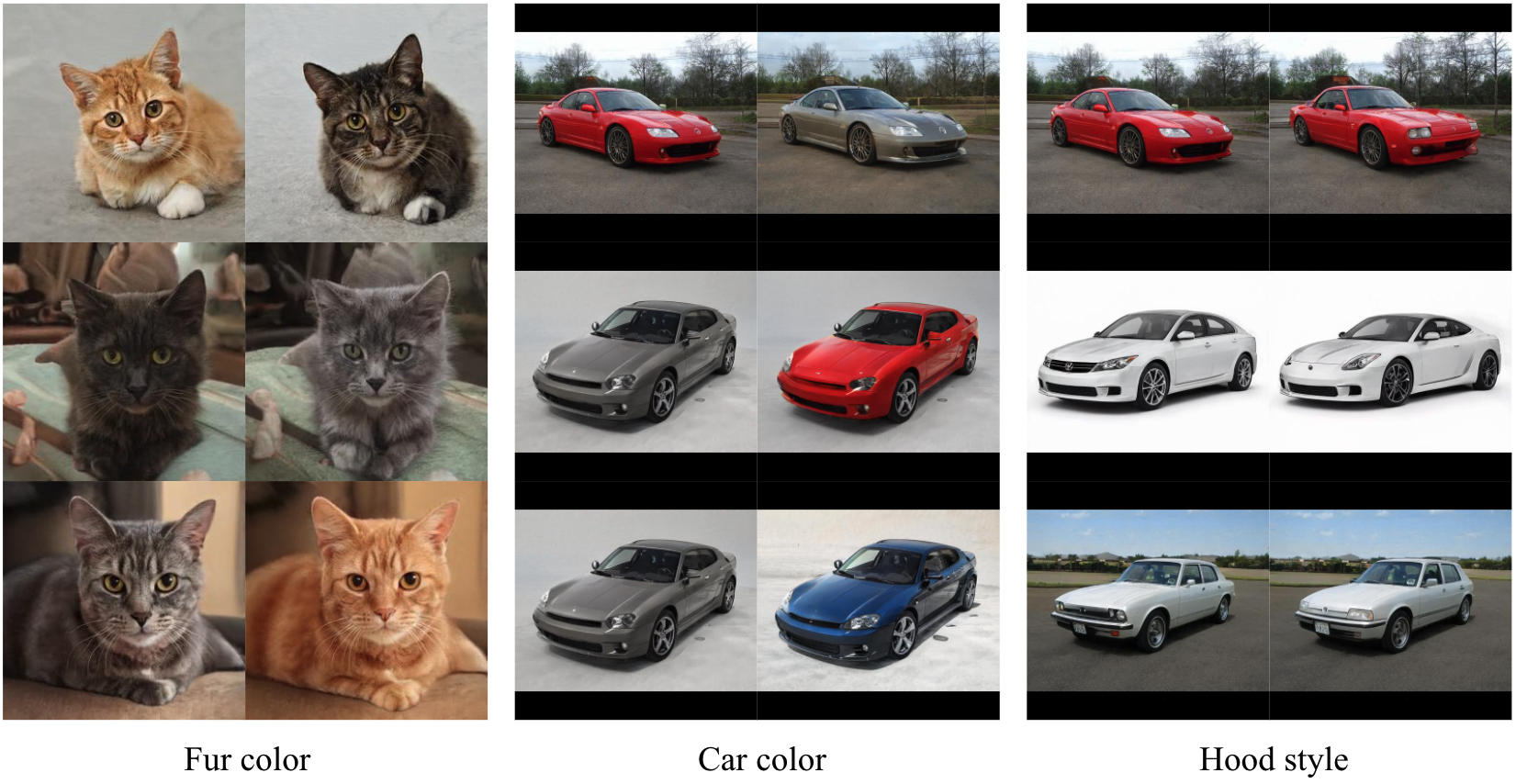}
\caption{Multi-class manipulation results for LSUN cats and cars~\cite{yu2015lsun} . For each group of images, left: source, right: modified. Regardless of the original color of the car/cat, our proposed method can independently identify any target control direction.}
\end{subfigure}
\begin{subfigure}{1\textwidth}

\includegraphics[width=\textwidth]{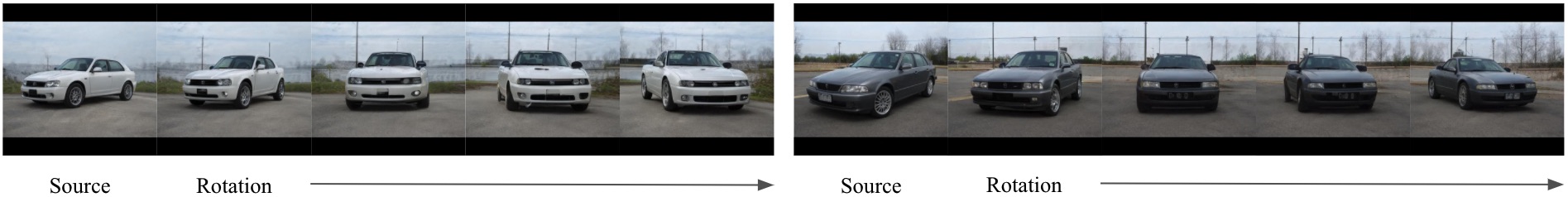}
\caption{Multi-class manipulation results for LSUN cars rotation. To avoid 3D-unaware changes, starting from the initial angle class, the latent code is optimized towards increments of 45 degrees which correspond to different classes.}

\end{subfigure}

\caption{Multi-class manipulation for LSUN cats and LSUN cars~\cite{yu2015lsun}. Our method can easily apply multi-directional changes without learning each direction between each pair of classes.}
\label{multi_class}
\end{figure}

\noindent
\textbf{Multiclass Manipulation.}
In this section, we showcase our method's ability to manipulate multiclass attributes.
Based on the latent-semantic Jacobian (Eqn.~\ref{eq:latent-semantic-jacobian}), regardless of the original state of the image, we identify the control direction and the activated channels with respect to the target class.
Using only one classifier, our method performs manipulations in any direction (Fig.~\ref{multi_class}).

\begin{figure}[tb]
\centering
\includegraphics[width=\textwidth]{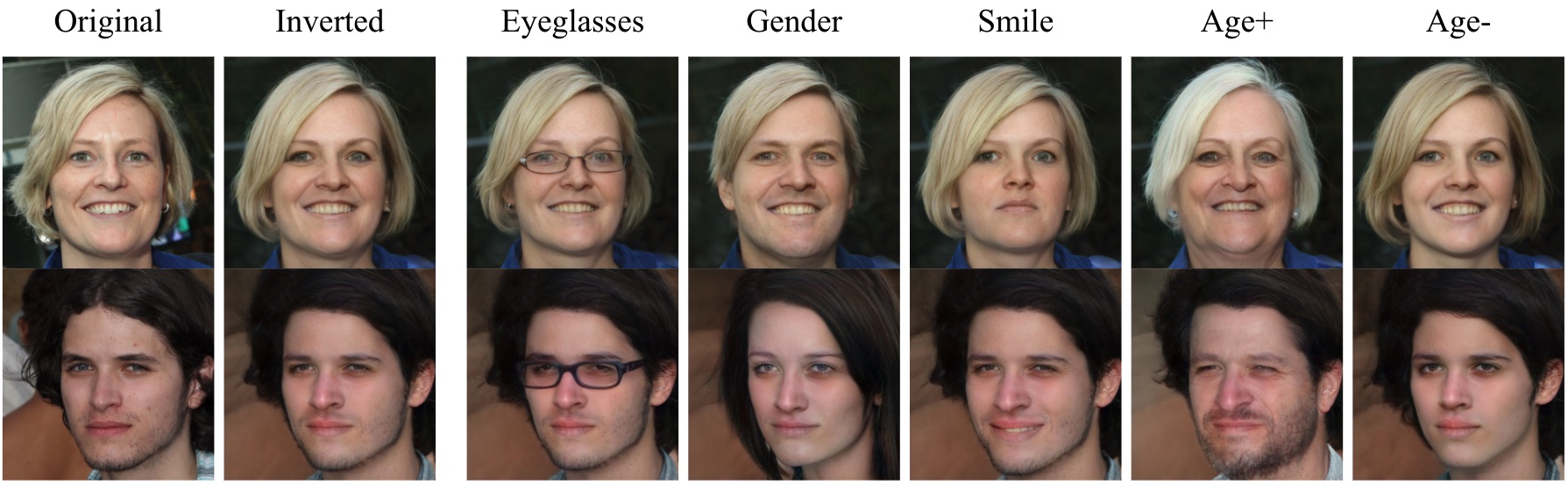}
\caption{Manipulating real faces by first inverting them into $W$ space.}
\label{inversion}

\end{figure}

\noindent
\textbf{Real Image Manipulation.}
We present real image editing results to verify that the semantic controls learned by our method can be applied to real images.
We first invert real human face images into the $W$ space of StyleGAN2~\cite{karras2020analyzing} trained on FFHQ~\cite{karras2019style} via an optimization-based framework~\cite{karras2020analyzing}.
We then interpolate the latent codes along directions learned by our method.
In the resulting manipulations (Fig.~\ref{inversion}), our method manipulates all attributes successfully.

\subsection{Quantitative Evaluation}
We compare our method with the related works~\cite{harkonen2020ganspace,shen2020interfacegan,wu2021stylespace} by measuring the accuracy and the level of attribute entanglement.
We test on StyleGAN2~\cite{karras2020analyzing} pretrained on the FFHQ~\cite{karras2019style} dataset.
We generate a test image bank with 500k samples using the StyleGAN2 model and score 4 attributes (age, gender, smile and eyeglasses) with the corresponding pre-trained CelebA classifiers~\cite{karras2019style}.
From our test image bank, we sample 3k images with logits around the decision boundaries for 4 attributes, and for each image, we sample a target attribute $k$ uniformly from the 4 to optimize until the modified attribute logit crosses the boundary, and we include the details regarding the algorithm in the Appendix.

In a nutshell, we manipulate the latent code until the target attribute classification changes. We stop after crossing the boundary as for attributes like smile and eyeglasses, further optimization results in regions in the latent space irrelevant to the original target, e.g., removing eyeglasses from a person not wearing eyeglasses generally makes the person younger.

\begin{table}[ht]

 \caption{Attribute manipulation accuracy in a disentangled manner. A resulting image is considered a true positive if only the target attribute is successfully changed. Among all, our method achieves the highest scores for all 4 attributes, corresponding to fewer changes in attributes other than the target during manipulation, and higher success rates, i.e., target attribute being edited successfully.}
\label{accuracy}
\begin{center}
\begin{tabular}{*{5}{c}}
\toprule
Accuracy & Gender & Smile & Eyeglasses & Age \\
\midrule
InterFaceGAN & 0.5859 & 0.9033 & 0.5946 & 0.5620 \\
GANSpace & 0.2770 & 0.6146 & x & 0.0063 \\
StyleSpace & 0.0812 & 0.7390 & 0.3510 & 0.0034 \\
\textbf{Ours} & \textbf{0.6937} & \textbf{0.9254} & \textbf{0.6626} & \textbf{0.6139} \\
\bottomrule
\end{tabular}
\end{center}

\end{table}
\begin{figure}[tb]
\centering
\begin{subfigure}{1\textwidth}
\centering

\includegraphics[width=\textwidth]{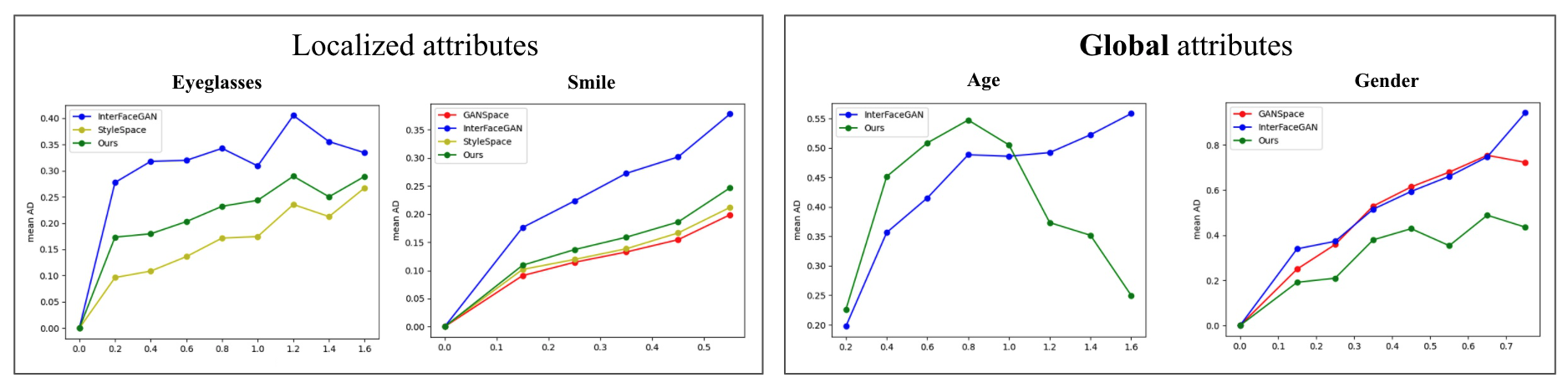}

\caption{X-axis: Change in logits of the target attribute. Y-axis: Mean AD. Lower AD indicates better disentanglement. Our method (green) is significantly less entangled than InterFaceGAN and GANSpace for gender, which involves global-level changes, and achieves similar levels of disentanglement for editing smile compared to StyleSpace and GANSpace. Note that unnatural looking and identity change will not affect this calculation.}
\label{ad}
\end{subfigure}
\begin{subfigure}{1\textwidth}
\centering
\includegraphics[width=\textwidth]{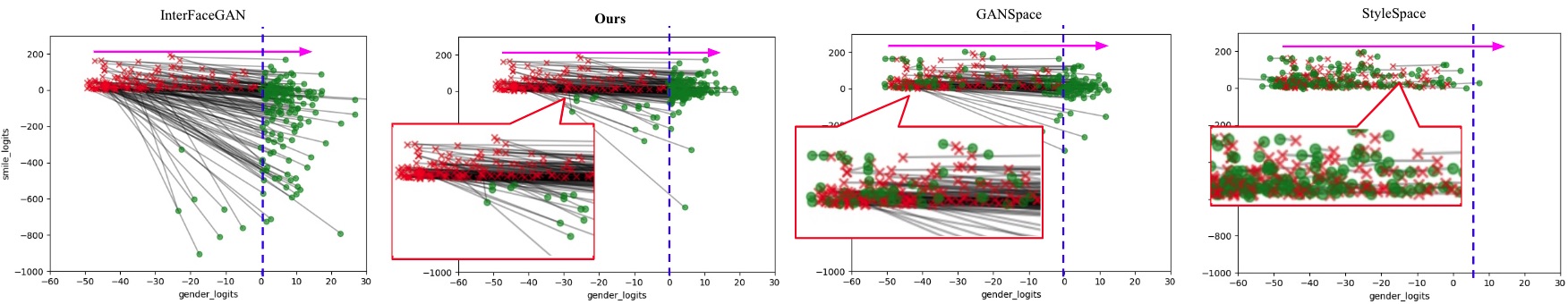}
\caption{Disentanglement visualization for gender and smile. X-axis: gender logits; Y-axis: smile logits. Red crosses: original logits; Green dots: logits after modification. Pink arrow: change in logits when attributes are perfectly disentangled. Blue dash line: success threshold. Flatter lines indicate less disentanglement. More green dots lying to the right of the blue dash line indicate a higher success rate. Ours exhibits the best disentanglement with a high success rate.}
\label{connection}
\end{subfigure}
\caption{Disentanglement analysis}
\label{disentanglement}

\end{figure}

\noindent
\textbf{Attribute Manipulation Accuracy.} We measure whether the attributes of an optimized image match the intended targets, i.e., change in the target attribute label and unchanged labels for the rest (Tab.~\ref{accuracy}).
High accuracy corresponds to a low level of attribute entanglement and high success rate, i.e., target attribute being present.
Our approach achieves the best results among all.

\noindent
\textbf{Attribute Disentanglement.}
Attribute Dependency (AD)~\cite{wu2021stylespace} measures the level of entanglements for semantic controls in GAN latent space, and here we follow a similar paradigm. On a high level, for images with target attribute $k \in A$ being manipulated, where $A$ stands for all attributes, we group the modified images by $x= \frac{\Delta l_{k}}{\sigma_k}$. We then compute mean-AD with $ E(\frac{1}{|A|-1} \Sigma_{i \in A\setminus k}\frac{\Delta l_i}{\sigma l_i})$ for each group. The full algorithm can be found in the appendix.

 (Fig.~\ref{ad}) plots the mean AD for each target attribute, where methods with extremely low accuracy are excluded, including StyleSpace for gender and age and GANSpace for eyeglasses as it doesn't offer such control. 
 For attributes that only require localized changes, including eyeglasses and smile, our method performs similarly to StyleSpace and GANSpace. 
 Our eyeglasses control shows higher AD than StyleSpace, as StyleSpace only modifies channels responsible for the eye area, yet for ours, the image is edited on a global level, and the pre-trained classifiers might be sensitive to some subtle changes.
Nevertheless, our method significantly outperforms the rest when editing gender and age, as the amount of change in the target increases. 
As we traverse longer distances in the latent space to increase a person's age, the linear directions learned by the SVMs become inaccurate with lots of entanglement with gender and eyeglasses, whereas our non-linear method continues to disentangle the target direction from the others. 

 To provide further insights for AD, we visualize the change in smile logits when changing from non-smiling male to female in (Fig.~\ref{connection}), and big negative slopes indicate entanglement between smile and gender, i.e., smile being added after optimization. 
InterFaceGAN suffers from the entanglement issue with a lot of big negative slopes, whereas GANSpace and StyleSpace have a lot of failure cases (the green points lying on the left of the blue vertical threshold line after interpolation in (Fig.~\ref{connection})), indicating manipulation in the wrong/unclear semantic direction. Our method has much less entanglement (slopes are more similar to the flat pink arrow) with fewer failure cases, and for those that failed, the direction is mostly correct as the gender logits are still moving towards the blue dash line. 

\begin{figure}[tb]
\centering
\begin{subfigure}{1\textwidth}
\centering
\includegraphics[width=\textwidth]{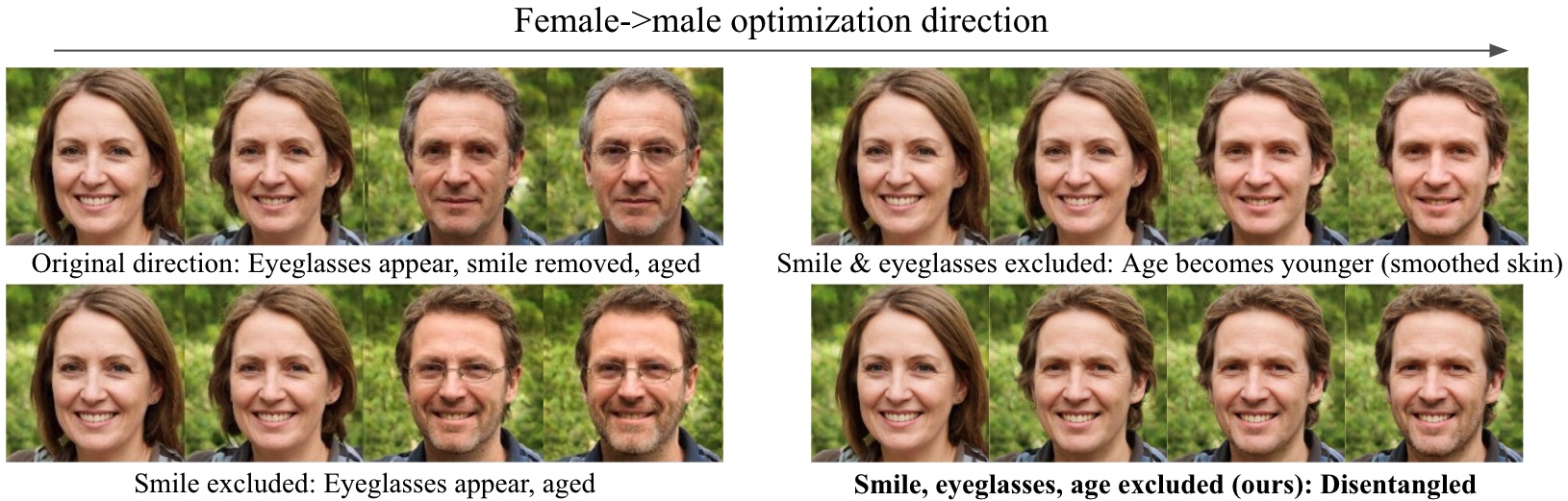}

\caption{Disentangle gender from other attributes one by one. The upper left row shows entanglement with eyeglasses, smile and age. At the bottom right row, by excluding the union of the activated channels for all entangled attributes, we successfully achieve disentanglement, where only the gender attribute is changed.}
\label{ablation_ffhq}
\end{subfigure}
\begin{subfigure}{1\textwidth}
\centering

\includegraphics[width=\textwidth]{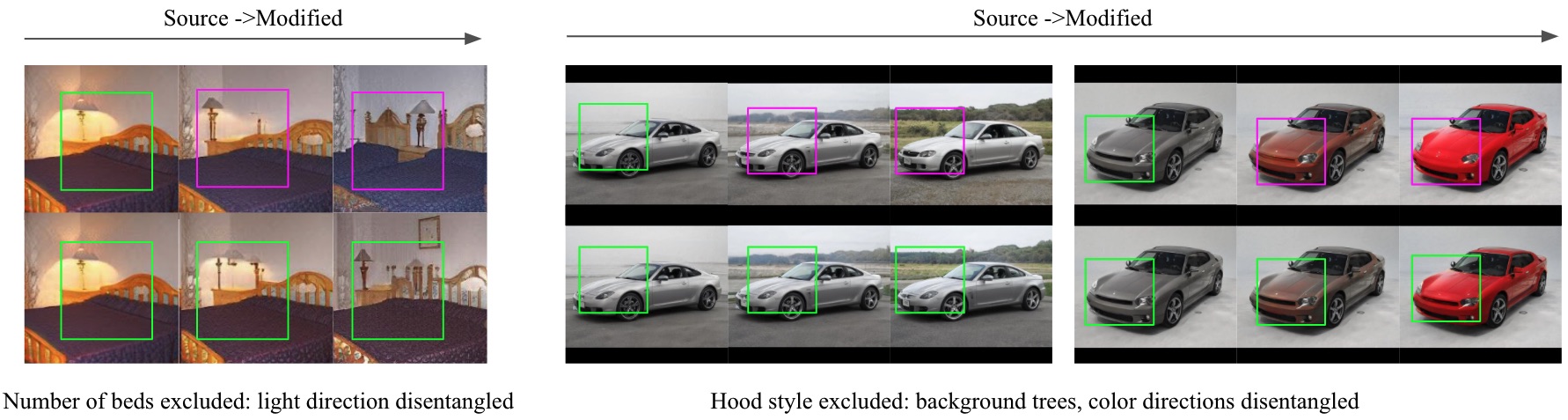}
\caption{Disentanglement results for LSUN-Cars and LSUN-Bedrooms~\cite{yu2015lsun}. The first row and second row shows original direction and disentangled direction respectively. }
\label{ablation_lsun}
\end{subfigure}

\begin{subfigure}{1\textwidth}
\centering

\includegraphics[width=\textwidth]{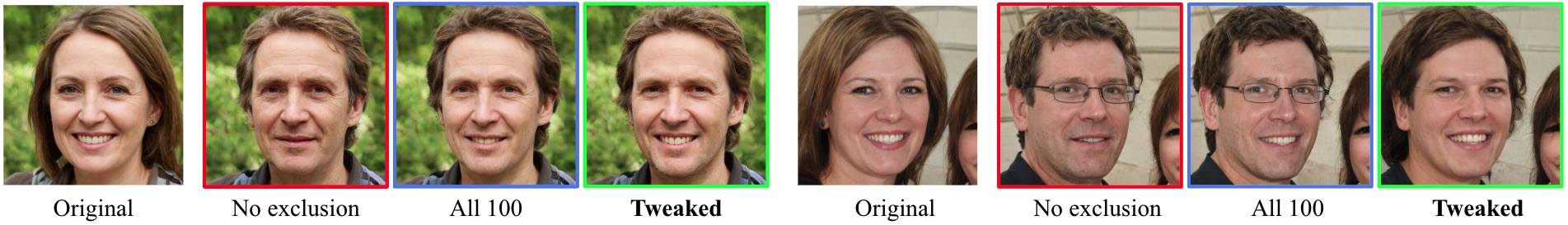}
\caption{Effect of number of channels excluded for smile, age, eyeglasses when editing gender. Tweaked: results shown in (Fig. ~\ref{ffhq_compare}).}
\label{threshold}
\end{subfigure}

\caption{Ablation studies on disentanglement.}

\end{figure}






\subsection{Ablation Studies: Attribute Disentanglement}
In this section, we demonstrate how controls over different target attributes computed by our method can be disentangled using gradient information from each entangled attribute. 
We first demonstrate how our method enables disentanglement from multiple attributes for StyleGAN2~\cite{karras2020analyzing} pre-trained on the FFHQ~\cite{karras2019style} dataset in (Fig.~\ref{ablation_ffhq}), where the original gender direction is entangled with smile, eyeglasses and age. 
By filtering out activated channels for the entangled attribute(s) calculated from the gradient weights in our classification network, we achieve different levels of disentanglement. 
We further present disentanglement results for StyleGAN2~\cite{karras2020analyzing} pretrained on the LSUN-Cars~\cite{yu2015lsun} dataset and StyleGAN~\cite{karras2019style} pretrained on the LSUN-Bedrooms dataset ~\cite{yu2015lsun} in (Fig.~\ref{ablation_lsun}).

Next, we study the \textbf{robustness} of attribute disentanglement with respect to the number of top-$k$ channels selected. 
When setting $k$ to 100 for all attributes as shown in (Fig.~\ref{threshold}), our method produces disentangled results, and the level of disentanglement for editing each attribute can be further improved by manually adjusting based on the results on a small meta-validation set of 30 examples with $k$ in \{50,100,150,200,250\}.



\section{Discussion}
%
%

\noindent
\textbf{Limitations.} Despite the overall success of our method, we also notice some failures which lack 3D understanding (e.g. car rotation) or result in identity change. Examples and discussions can be found in the Appendix. 

\noindent
\textbf{Conlusions.} In this work, we propose a simple but powerful approach based on gradients to obtain multi-directional controls and disentangle one attribute from the others. 
Our findings show that attributes' controls can be separated even with strongly correlated training datasets via gradient information.
This work shows considerable potential to discover rich controls in the learned latent space via gradient-based information, as well as its effectiveness in small data regimes. 


\clearpage
%
%
\bibliographystyle{splncs04}
\bibliography{egbib}

\begin{thebibliography}{10}
\providecommand{\url}[1]{\texttt{#1}}
\providecommand{\urlprefix}{URL }
\providecommand{\doi}[1]{https://doi.org/#1}

\bibitem{bau2019gandissect}
Bau, D., Zhu, J.Y., Strobelt, H., Zhou, B., Tenenbaum, J.B., Freeman, W.T.,
  Torralba, A.: Gan dissection: Visualizing and understanding generative
  adversarial networks. In: Proceedings of the International Conference on
  Learning Representations (ICLR) (2019)

\bibitem{bau2019seeing}
Bau, D., Zhu, J.Y., Wulff, J., Peebles, W., Strobelt, H., Zhou, B., Torralba,
  A.: Seeing what a gan cannot generate. In: Proceedings of the IEEE/CVF
  International Conference on Computer Vision. pp. 4502--4511 (2019)

\bibitem{brock2018large}
Brock, A., Donahue, J., Simonyan, K.: Large scale {GAN} training for high
  fidelity natural image synthesis. In: International Conference on Learning
  Representations (2019), \url{https://openreview.net/forum?id=B1xsqj09Fm}

\bibitem{chattopadhay2018grad}
Chattopadhay, A., Sarkar, A., Howlader, P., Balasubramanian, V.N.: Grad-cam++:
  Generalized gradient-based visual explanations for deep convolutional
  networks. In: 2018 IEEE winter conference on applications of computer vision
  (WACV). pp. 839--847. IEEE (2018)

\bibitem{chen2021learning}
Chen, Y., Liu, S., Wang, X.: Learning continuous image representation with
  local implicit image function. In: Proceedings of the IEEE/CVF Conference on
  Computer Vision and Pattern Recognition. pp. 8628--8638 (2021)

\bibitem{cherepkov2021navigating}
Cherepkov, A., Voynov, A., Babenko, A.: Navigating the gan parameter space for
  semantic image editing. In: Proceedings of the IEEE/CVF Conference on
  Computer Vision and Pattern Recognition. pp. 3671--3680 (2021)

\bibitem{choi2018stargan}
Choi, Y., Choi, M., Kim, M., Ha, J.W., Kim, S., Choo, J.: Stargan: Unified
  generative adversarial networks for multi-domain image-to-image translation.
  In: Proceedings of the IEEE conference on computer vision and pattern
  recognition. pp. 8789--8797 (2018)

\bibitem{choi2020stargan}
Choi, Y., Uh, Y., Yoo, J., Ha, J.W.: Stargan v2: Diverse image synthesis for
  multiple domains. In: Proceedings of the IEEE/CVF conference on computer
  vision and pattern recognition. pp. 8188--8197 (2020)

\bibitem{goodfellow2014generative}
Goodfellow, I., Pouget-Abadie, J., Mirza, M., Xu, B., Warde-Farley, D., Ozair,
  S., Courville, A., Bengio, Y.: Generative adversarial nets. Advances in
  neural information processing systems  \textbf{27} (2014)

\bibitem{harkonen2020ganspace}
H{\"a}rk{\"o}nen, E., Hertzmann, A., Lehtinen, J., Paris, S.: Ganspace:
  Discovering interpretable gan controls. Advances in Neural Information
  Processing Systems  \textbf{33},  9841--9850 (2020)

\bibitem{Isola_2017_CVPR}
Isola, P., Zhu, J.Y., Zhou, T., Efros, A.A.: Image-to-image translation with
  conditional adversarial networks. In: Proceedings of the IEEE Conference on
  Computer Vision and Pattern Recognition (CVPR) (July 2017)

\bibitem{karras2019style}
Karras, T., Laine, S., Aila, T.: A style-based generator architecture for
  generative adversarial networks. In: Proceedings of the IEEE/CVF conference
  on computer vision and pattern recognition. pp. 4401--4410 (2019)

\bibitem{karras2020analyzing}
Karras, T., Laine, S., Aittala, M., Hellsten, J., Lehtinen, J., Aila, T.:
  Analyzing and improving the image quality of stylegan. In: Proceedings of the
  IEEE/CVF conference on computer vision and pattern recognition. pp.
  8110--8119 (2020)

\bibitem{ledig2017photo}
Ledig, C., Theis, L., Husz{\'a}r, F., Caballero, J., Cunningham, A., Acosta,
  A., Aitken, A., Tejani, A., Totz, J., Wang, Z., et~al.: Photo-realistic
  single image super-resolution using a generative adversarial network. In:
  Proceedings of the IEEE conference on computer vision and pattern
  recognition. pp. 4681--4690 (2017)

\bibitem{li2021continuous}
Li, Z., Jiang, R., Aarabi, P.: Continuous face aging via self-estimated
  residual age embedding. In: Proceedings of the IEEE/CVF Conference on
  Computer Vision and Pattern Recognition. pp. 15008--15017 (2021)

\bibitem{liu2019stgan}
Liu, M., Ding, Y., Xia, M., Liu, X., Ding, E., Zuo, W., Wen, S.: Stgan: A
  unified selective transfer network for arbitrary image attribute editing. In:
  Proceedings of the IEEE conference on computer vision and pattern
  recognition. pp. 3673--3682 (2019)

\bibitem{park2019semantic}
Park, T., Liu, M.Y., Wang, T.C., Zhu, J.Y.: Semantic image synthesis with
  spatially-adaptive normalization. In: Proceedings of the IEEE/CVF conference
  on computer vision and pattern recognition. pp. 2337--2346 (2019)

\bibitem{radford2015unsupervised}
Radford, A., Metz, L., Chintala, S.: Unsupervised representation learning with
  deep convolutional generative adversarial networks (2016)

\bibitem{saha2021LOHO}
Saha, R., Duke, B., Shkurti, F., Taylor, G., Aarabi, P.: Loho: Latent
  optimization of hairstyles via orthogonalization. In: CVPR (2021)

\bibitem{selvaraju2017grad}
Selvaraju, R.R., Cogswell, M., Das, A., Vedantam, R., Parikh, D., Batra, D.:
  Grad-cam: Visual explanations from deep networks via gradient-based
  localization. In: Proceedings of the IEEE international conference on
  computer vision. pp. 618--626 (2017)

\bibitem{shen2020interfacegan}
Shen, Y., Yang, C., Tang, X., Zhou, B.: Interfacegan: Interpreting the
  disentangled face representation learned by gans. IEEE transactions on
  pattern analysis and machine intelligence  (2020)

\bibitem{smilkov2017smoothgrad}
Smilkov, D., Thorat, N., Kim, B., Vi{\'e}gas, F., Wattenberg, M.: Smoothgrad:
  removing noise by adding noise. arXiv preprint arXiv:1706.03825  (2017)

\bibitem{sundararajan2017axiomatic}
Sundararajan, M., Taly, A., Yan, Q.: Axiomatic attribution for deep networks.
  In: International conference on machine learning. pp. 3319--3328. PMLR (2017)

\bibitem{upchurch2017deep}
Upchurch, P., Gardner, J., Pleiss, G., Pless, R., Snavely, N., Bala, K.,
  Weinberger, K.: Deep feature interpolation for image content changes. In:
  Proceedings of the IEEE conference on computer vision and pattern
  recognition. pp. 7064--7073 (2017)

\bibitem{voynov2020unsupervised}
Voynov, A., Babenko, A.: Unsupervised discovery of interpretable directions in
  the gan latent space. In: International Conference on Machine Learning. pp.
  9786--9796. PMLR (2020)

\bibitem{wang2018high}
Wang, T.C., Liu, M.Y., Zhu, J.Y., Tao, A., Kautz, J., Catanzaro, B.:
  High-resolution image synthesis and semantic manipulation with conditional
  gans. In: Proceedings of the IEEE conference on computer vision and pattern
  recognition. pp. 8798--8807 (2018)

\bibitem{wu2019logan}
Wu, Y., Donahue, J., Balduzzi, D., Simonyan, K., Lillicrap, T.: Logan: Latent
  optimisation for generative adversarial networks. arXiv preprint
  arXiv:1912.00953  (2019)

\bibitem{wu2021stylespace}
Wu, Z., Lischinski, D., Shechtman, E.: Stylespace analysis: Disentangled
  controls for stylegan image generation. In: Proceedings of the IEEE/CVF
  Conference on Computer Vision and Pattern Recognition. pp. 12863--12872
  (2021)

\bibitem{yang2021semantic}
Yang, C., Shen, Y., Zhou, B.: Semantic hierarchy emerges in deep generative
  representations for scene synthesis. International Journal of Computer Vision
   \textbf{129}(5),  1451--1466 (2021)

\bibitem{yeh2017semantic}
Yeh, R.A., Chen, C., Yian~Lim, T., Schwing, A.G., Hasegawa-Johnson, M., Do,
  M.N.: Semantic image inpainting with deep generative models. In: Proceedings
  of the IEEE conference on computer vision and pattern recognition. pp.
  5485--5493 (2017)

\bibitem{yu2015lsun}
Yu, F., Seff, A., Zhang, Y., Song, S., Funkhouser, T., Xiao, J.: Lsun:
  Construction of a large-scale image dataset using deep learning with humans
  in the loop. arXiv preprint arXiv:1506.03365  (2015)

\bibitem{yu2019free}
Yu, J., Lin, Z., Yang, J., Shen, X., Lu, X., Huang, T.S.: Free-form image
  inpainting with gated convolution. In: Proceedings of the IEEE/CVF
  International Conference on Computer Vision. pp. 4471--4480 (2019)

\bibitem{zhu2017unpaired}
Zhu, J.Y., Park, T., Isola, P., Efros, A.A.: Unpaired image-to-image
  translation using cycle-consistent adversarial networks. In: Proceedings of
  the IEEE international conference on computer vision. pp. 2223--2232 (2017)

\end{thebibliography}
\end{document}